\documentclass[conference]{IEEEtran}
\IEEEoverridecommandlockouts
\usepackage{cite}
\usepackage{amsmath,amssymb,amsfonts}
\usepackage{graphicx}
\usepackage{textcomp}
\usepackage{xcolor}
\usepackage{booktabs}
\usepackage{array}
\usepackage{multirow}
\usepackage{url}

\graphicspath{{./}}

\def\BibTeX{{\rm B\kern-.05em{\sc i\kern-.025em b}\kern-.08em
    T\kern-.1667em\lower.7ex\hbox{E}\kern-.125emX}}

\begin{document}

\title{Efficient LLM Distillation for Bangladesh Legal Context: 
A Smartphone-Compatible Retrieval-Augmented Generation Model}

\author{
\IEEEauthorblockN{}
\IEEEauthorblockA{
\begin{tabular}{c@{\hspace{2.5em}}c@{\hspace{2.5em}}c}
\begin{tabular}[t]{@{}c@{}}
\textbf{MD. Nafis Kamal}\\
\textit{Computer Science and Engineering}\\
\textit{BRAC University}\\
Dhaka, Bangladesh\\
md.nafis.kamal@g.bracu.ac.bd
\end{tabular}
&
\begin{tabular}[t]{@{}c@{}}
\textbf{Mahadi Hasan Fahim}\\
\textit{Computer Science}\\
\textit{BRAC University}\\
Dhaka, Bangladesh\\
mahadi.hasan.fahim@g.bracu.ac.bd
\end{tabular}
&
\begin{tabular}[t]{@{}c@{}}
\textbf{Talha Ridwan}\\
\textit{Computer Science}\\
\textit{BRAC University}\\
Dhaka, Bangladesh\\
talha.ridwan@g.bracu.ac.bd
\end{tabular}
\\[8em]
\begin{tabular}[t]{@{}c@{}}
\textbf{Nadifa Zaman}\\
\textit{Computer Science}\\
\textit{BRAC University}\\
Dhaka, Bangladesh\\
nadifa.zaman@g.bracu.ac.bd
\end{tabular}
&
\begin{tabular}[t]{@{}c@{}}
\textbf{Fariha Roushon Florin}\\
\textit{Computer Science}\\
\textit{BRAC University}\\
Dhaka, Bangladesh\\
fariha.roushon.florin@g.bracu.ac.bd
\end{tabular}
&
\begin{tabular}[t]{@{}c@{}}
\textbf{Dr. Farig Yousuf Sadeque}\\
\textit{Associate Professor}\\
\textit{Computer Science and Engineering}\\
\textit{BRAC University}\\
Dhaka, Bangladesh\\
farig.sadeque@bracu.ac.bd
\end{tabular}
\\[8em]
\multicolumn{3}{c}{
\begin{tabular}[t]{@{}c@{}}
\textbf{Saadat Rafid Ahmed}\\
\textit{Lecturer}\\
\textit{Computer Science and Engineering}\\
\textit{BRAC University}\\
Dhaka, Bangladesh\\
saadat.ahmed@bracu.ac.bd
\end{tabular}
}
\end{tabular}
}
}

\maketitle

\begin{abstract}
Legal information in Bangladesh is inaccessible to most citizens. Statutory text is English-only, trained lawyers are concentrated in urban centres, and cloud-dependent AI fails where mobile connectivity is unreliable, a setting in which hallucinated legal text causes direct harm.
The system addresses statutory interpretation only; queries that require judicial precedent or case-law reasoning fall outside its scope.
We target the statutory access gap by compressing a 9-billion-parameter Gemma-2 teacher into a 2-billion-parameter student through two-phase progressive knowledge distillation. Phase~1 performs supervised fine-tuning on 9,429 quality-gated legal question--answer pairs (65\% acceptance from 14,514 generated queries); Phase~2 minimises sparse Kullback--Leibler divergence against the teacher's top-50 per-token logits at temperature $\tau{=}4.0$, implemented via QLoRA (4-bit NF4, rank-32 LoRA adapters).
Prior legal language models target general legal English; this system specialises in Bangladeshi statutory law and operates without network access.
Every response is grounded through hybrid retrieval that combines dense semantic search (60\%) and BM25 (40\%) across 36,029 statutory passages from the Bangladesh Constitution and national legislation.
On a 50-query English benchmark, the distilled model reaches ROUGE-L~0.4715 and BERTScore~F1~0.5679, a 103\% ROUGE-L and 143\% BERTScore gain over the retrieval-augmented undistilled baseline (ROUGE-L~0.2323, BERTScore~0.2340).
The adapter quantises to 1.6~GB (GGUF Q4\_K\_M) and runs at 4--8~tokens per second on a Pixel~6 without network access.
Cross-lingual evaluation on 50~Bangla queries yields ROUGE-L~0.4083 and BERTScore~0.8133, showing effective retrieval from Bangla input against an English-only corpus.
In a single-evaluator pilot, a practising lawyer rated 50~responses at a weighted mean of 4.16/5 (90\% rated~4 or~5), which supports practical utility beyond text-overlap metrics.%
\end{abstract}


\section{Introduction}

Large language models help with legal research. They summarise statutes, flag relevant provisions, and draft preliminary analyses.
Their failure mode is just as well known: they hallucinate, producing confident but factually wrong legal claims~\cite{terzidou2025generative,Li2024}.
In most fields a hallucination is an embarrassment. In legal advice it can cost someone their case or their liberty.

Bangladesh makes this harder for three specific reasons.
First, the legal corpus is split. The Constitution and core national statutes are in English, while procedural and administrative instruments are in Bangla, and that bilingual divide defeats monolingual retrieval systems.
Second, cloud-dependent AI assistants are out of reach for most of the population, who rely on mobile devices over intermittent or metered connections~\cite{fan2024surveyragmeetingllms}.
Third, no legal AI exists for this jurisdiction. Encoder-only compressed models such as DistilBERT~\cite{sanh2020distilbertdistilledversionbert} and TinyBERT~\cite{jiao2020tinybertdistillingbertnatural} cannot generate answers, and Bangla-capable models such as BanglaBERT~\cite{bhattacharjee2022banglabertlanguagemodelpretraining} have neither legal domain adaptation nor generative capability.

Two techniques address different parts of the problem.
Retrieval-Augmented Generation (RAG)~\cite{lewis2021retrievalaugmentedgenerationknowledgeintensivenlp} grounds generation in retrieved statutory passages, which cuts hallucination risk. But standard sparse and dense retrievers struggle on legal text that mixes normalised citation identifiers with dense procedural language~\cite{li2025lexragbenchmarkingretrievalaugmentedgeneration,rayo2025hybridapproachinformationretrieval}.
Knowledge distillation compresses capable teachers into deployable students. Prior generative KD for legal tasks has used response-level supervision (SeqKD), copying teacher output text while discarding the per-token probability distributions that encode the teacher's uncertainty and ranked alternatives across legal interpretations~\cite{hinton2015distillingknowledgeneural,song2026surveyonpolicydistillationlarge}.

This paper closes both gaps for a single underserved jurisdiction.
The contributions are:
\begin{enumerate}
    \item A two-phase progressive distillation pipeline: Phase~1 fine-tunes on 9,429 quality-gated legal QA pairs (65\% acceptance from 14,514 generated queries); Phase~2 minimises sparse KL divergence against the teacher's top-50 per-token logits at temperature $\tau{=}4.0$ via QLoRA (4-bit NF4, rank-32 LoRA adapters), compressing a 9B teacher to a 2B student.
    \item A hybrid retrieval stage combining 60\% dense semantic search and 40\% BM25 over 36,029 statutory passages from the Bangladesh Constitution and national legislation, with cross-encoder re-ranking.
    \item A complete mobile deployment pipeline producing a 1.6~GB GGUF Q4\_K\_M file that runs fully offline at 4--8~tokens per second on a Pixel~6.
    \item Evaluation showing 103\% ROUGE-L and 143\% BERTScore improvement over the retrieval-augmented undistilled baseline, cross-lingual Bangla results (ROUGE-L~0.4083, BERTScore~0.8133 with RAG), and practitioner validation by a qualified lawyer (weighted mean 4.16/5, $n{=}50$).
\end{enumerate}

\section{Related Work}

\subsection{LLMs and Hallucination in Legal AI}

The main risk of deploying general-purpose LLMs for legal tasks is confident wrongness rather than ignorance.
Models hallucinate authoritative-sounding case citations, statutory provisions, and procedural rules that do not exist; in legal advice, a single fabricated citation or misapplied provision can cost someone a claim or a case~\cite{terzidou2025generative,Li2024}.
Bangladeshi statutory law is particularly exposed: the corpus is largely absent from standard pre-training datasets, so base models have no reliable parametric knowledge to fall back on when retrieval context is missing.

\subsection{Retrieval-Augmented Generation for Legal Domains}

RAG~\cite{lewis2021retrievalaugmentedgenerationknowledgeintensivenlp} addresses hallucination at inference time by inserting retrieved source passages into the prompt, which constrains generation to verified statutory text.
Legal corpora impose retrieval challenges that general-domain pipelines were not designed for. Exact statutory identifiers such as section numbers, act names, and cross-references require lexical matching, while legal arguments require semantic generalisation, and both dense-only and sparse-only retrievers handle this split poorly~\cite{rayo2025hybridapproachinformationretrieval,wang2025balancingblendexperimentalanalysis}.
Multi-turn legal consultation adds pronoun resolution and context-accumulation requirements~\cite{li2025lexragbenchmarkingretrievalaugmentedgeneration}; citation-aware retrieval must resolve embedded cross-act references such as ``as defined in section~4 of [another Act]''~\cite{panchal2025lawpalretrievalaugmented}.
None of these systems address Bangladeshi law, offline operation, or mobile-constrained hardware.

\subsection{Knowledge Distillation for Generative Models}

Hinton et al.~\cite{hinton2015distillingknowledgeneural} showed that training a student on the teacher's full softmax distribution at temperature $\tau > 1$ transfers richer information than one-hot labels: the soft targets expose the teacher's uncertainty and relative confidence across near-equivalent outputs, which is the distributional signal a model needs for nuanced legal reasoning.
For encoder-only models this produced DistilBERT~\cite{sanh2020distilbertdistilledversionbert}, TinyBERT~\cite{jiao2020tinybertdistillingbertnatural}, and MobileBERT~\cite{sun2020mobilebertcompacttaskagnosticbert}.
Extending it to auto-regressive LLMs is harder. Sequence-level KD copies the teacher's decoded output text but collapses per-token distributions to a single mode, discarding uncertainty~\cite{song2026surveyonpolicydistillationlarge,borkar2026memorizationdynamicsknowledgedistillation}.
Prior generative legal LLMs have relied on instruction fine-tuning rather than logit-matching distillation, and target English common-law jurisdictions that require cloud inference.
QLoRA~\cite{dettmers2023qloraefficientfinetuningquantized} enables LoRA adaptation on 4-bit NF4 bases, bringing billion-parameter domain fine-tuning within reach of a single consumer GPU.

\subsection{Mobile LLM Deployment}

Edge deployment requires shrinking both model footprint and memory bandwidth without erasing task capability.
GGUF Q4\_K\_M quantisation reduces file size by 60--70\% relative to float16 across the 2--9B parameter range with modest quality degradation~\cite{kurt2026quantizationiuseunified}; the 2B student's merged float16 checkpoint (5.23~GB) quantises to 1.6~GB in our setting.
The GGUF format and \texttt{llama.cpp} runtime deliver CPU inference through SIMD-vectorised kernels and memory-mapped weights~\cite{llamacpp2023,prieto2025edgedeploymentsmalllanguage,tummalapalli2026llminferenceedgemobile}.
Prior on-device LLM work targets general-purpose assistants; no published work combines distilled domain-specific legal models with hybrid on-device retrieval in a fully offline Android deployment.

\section{System Architecture}

\begin{figure*}[t]
    \centering
    \includegraphics[width=\textwidth]{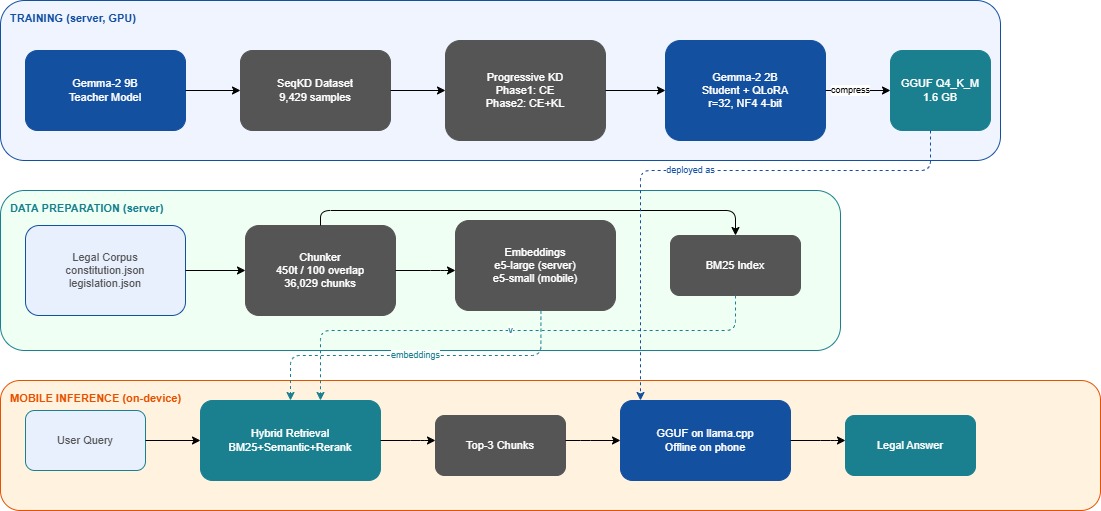}
    \caption{End-to-end pipeline: legal corpus preparation, progressive knowledge distillation from 9B teacher to 2B student, hybrid RAG retrieval, and offline mobile GGUF deployment.}
    \label{fig:system_overview}
\end{figure*}

The system has four interdependent components, each targeting a distinct failure mode of general-purpose LLMs in the Bangladeshi legal setting (Fig.~\ref{fig:system_overview}).

A \textit{corpus preparation module} crawls the Bangladesh Law and Justice Division repository, converts 1,083~legislative acts and 170~constitutional articles into 36,029~token-bounded statutory passages, and builds both semantic-embedding and BM25 indices to support hybrid retrieval downstream.

The \textit{progressive distillation pipeline} generates 9,429~quality-gated legal QA pairs using the 9B teacher, captures top-50 per-token logits at each output position, and trains the 2B student in two phases: cross-entropy alignment first, then sparse KL divergence against the stored distributions.

At inference, the \textit{hybrid RAG module} expands the user query into three paraphrases, scores all 36,029~passages with a 60/40 semantic-BM25 combination, filters by cosine threshold, and re-ranks the top-10 candidates with a cross-encoder before injecting the top-3 into the generation prompt.

The \textit{mobile deployment module} merges the trained adapter, converts to GGUF Q4\_K\_M (1.6~GB), and packages the model alongside a pre-computed embedding matrix and full corpus index for fully offline Android inference.

\section{Methodology}

\subsection{Legal Corpus Preparation}

We drew the corpus from the Bangladesh Law and Justice Division's public repository at \url{http://bdlaws.minlaw.gov.bd}.
HTML cleaning and quality filtering removed repealed provisions, malformed records, and placeholder text, since that material would present legally void content as current law if retrieved.
After filtering, 1,083~clean legislative acts spanning 1799--2025 and 170~constitutional articles remained.
The 226-year span is deliberate. Pre-independence colonial-era statutes are still operative in Bangladesh's legal system, so including them keeps the corpus aligned with the full body of applicable law rather than only post-independence legislation.

We chunked documents with a token-aware sentence-boundary splitter using a 450-token ceiling and 100-token overlap, which produced 36,029~passages stored in Parquet format with per-chunk metadata (act title, section type, token count, language).
We calibrated the ceiling by ablation. Bangladeshi legal provisions usually run 300--500 tokens, and a 256-token limit kept cutting provisions in half mid-clause, which dropped ROUGE-L by 10.1\% (Table~\ref{tab:chunk_ablation}).
Swapping the simpler earlier splitter (which had yielded $\sim$8,000 chunks) for the sentence-boundary-aware version recovered another 0.012 ROUGE-L (0.4595~$\to$~0.4715) by keeping more complete legal provisions per chunk.
We built three embedding indices from the same 36,029 chunks (Table~\ref{tab:corpus_stats}): \texttt{intfloat/multilingual-e5-large} (1024-dim) for server evaluation and SeqKD generation; \texttt{intfloat/multilingual-e5-small} (384-dim) as a conservative PC proxy for the GGUF mobile benchmark; and an ONNX-exported e5-large for on-device query-time retrieval.
All three map Bangla and English into a shared embedding space, so cross-lingual retrieval works without language detection.

\begin{table}[t]
\centering
\caption{Legal corpus statistics and embedding index configuration.}
\label{tab:corpus_stats}
\resizebox{\columnwidth}{!}{%
\begin{tabular}{@{}ll@{}}
\toprule
\textbf{Parameter} & \textbf{Value} \\
\midrule
Source documents & 1,083 acts + 170 constitutional articles \\
Chunk size (tokens) & 450 \\
Chunk overlap (tokens) & 100 \\
Total chunks & 36,029 \\
\midrule
\multicolumn{2}{@{}l}{\textit{Index 1 - server evaluation \& SeqKD generation}} \\
Embedding model & multilingual-e5-large (1024-dim) \\
Output file & \texttt{legal\_embeddings.parquet} ($\sim$148~MB) \\
\midrule
\multicolumn{2}{@{}l}{\textit{Index 2 - mobile GGUF benchmark (PC proxy)}} \\
Embedding model & multilingual-e5-small (384-dim) \\
Output file & \texttt{legal\_embeddings\_mobile.parquet} ($\sim$67~MB) \\
\midrule
\multicolumn{2}{@{}l}{\textit{Index 3 - on-device app (query-time retrieval)}} \\
Embedding model & multilingual-e5-large (1024-dim, ONNX) \\
Output file & \texttt{embeddings.bin} ($\sim$141~MB, memory-mapped) \\
\bottomrule
\end{tabular}%
}
\end{table}

\subsubsection*{Corpus Characteristics}

Three properties of the corpus shaped retrieval design decisions.

\textbf{Bilingual split.}
Bengali accounts for 56.4\% of legislative topics and English for 39.8\%, with 3.8\% mixed or unidentified.
This is not a symmetric balance: the English subset contains most of the exact statutory identifiers, including section numbers, act short-titles, and cross-references, while Bangla dominates procedural and administrative instruments.
A dense-only retriever handles Bangla queries well but tends to miss these English identifiers, and the 40\% BM25 component exists specifically to recover them.

\textbf{Boilerplate density.}
The five most common section headings are \textit{Definitions} (378~occurrences), \textit{Short Title and Commencement} (313), \textit{Power to Make Rules} (270), \textit{Repeal and Savings} (191), and \textit{Penalties} (187).
These recurring boilerplate sections inflate lexical overlap between unrelated acts, which makes keyword-only retrieval prone to returning procedurally similar but topically irrelevant chunks. The semantic component addresses this failure mode.

\textbf{Legislative distribution.}
Legislation is not uniformly distributed across time: the periods 1980--2009 (27.7\%) and 2010--2039 (27.3\%) together account for over half of all dated acts, which reflects Bangladesh's rapid post-independence legislative activity.
Pre-1950 acts are underrepresented because many were removed as repealed during cleaning rather than omitted, which is legally appropriate, since repealed provisions should not be returned as current law.

\subsection{Distillation Dataset Generation}

Dataset generation required two models working in sequence.
Gemma-2~27B via Ollama (\texttt{gemma2:27b}) first generated three synthetic legal queries per corpus chunk, skipping chunks under 30~words, producing 14,514~candidate queries from 4,838~processed chunks.
For each query, the teacher \texttt{google/gemma-2-9b-it} (loaded in 4-bit NF4) retrieved the top-3~relevant passages and produced a structured chain-of-thought response: (i)~identify the applicable legal principle, (ii)~apply it to the query, (iii)~state the legal conclusion, with intermediate reasoning enclosed in \texttt{<thought>} tags.

A quality gate filtered responses shorter than 80~characters or containing evasion phrases such as ``does not provide.'' These patterns signal that the retrieved statutory context lacked the specificity needed to ground a legally complete answer.
Of 14,514~queries, 9,429~passed (65\% acceptance). The rejected 35\% were structurally problematic queries whose corpus context did not support an answerable legal question rather than random failures, so their exclusion is deliberate.

Alongside each accepted response, the top-50 per-token probability distributions were captured at softening temperature $\tau=4.0$ and stored as \texttt{distillation\_logits\_top50\_9b.pt} ($\sim$1.9~GB, int32~indices~+~float16~probabilities).
At $\tau=4.0$ the top-50 tokens account for over 95\% of the teacher's probability mass across most output positions; storing the full 256,000-token vocabulary per position would require $\sim$30~GB for this dataset.
A 25-sample checkpoint interval prevented data loss across the multi-day generation run required to capture logits for all 9,429~samples.
Table~\ref{tab:dataset_stats} summarises the full configuration.

\begin{table}[t]
\centering
\caption{Distillation dataset generation statistics.}
\label{tab:dataset_stats}
\begin{tabular}{@{}ll@{}}
\toprule
\textbf{Parameter} & \textbf{Value} \\
\midrule
Teacher model & \texttt{google/gemma-2-9b-it} (4-bit NF4) \\
Query generation model & Gemma-2~27B via Ollama (\texttt{gemma2:27b}) \\
Total queries processed & 14,514 \\
Accepted samples & 9,429 \\
Acceptance rate & $\approx$65\% \\
Rejection criteria & Evasion phrases or ${<}$80 characters \\
Max new tokens (teacher) & 768 \\
Distillation temperature & 4.0 \\
Top-K logits saved per token & 50 \\
Logits storage format & int32 indices + float16 probabilities \\
Logits file size & $\sim$1.9~GB \\
Checkpoint interval & Every 25 samples \\
\bottomrule
\end{tabular}
\end{table}

\subsection{Progressive Knowledge Distillation}
\label{subsec:prog_kd}

\subsubsection*{Motivation from Failed Baselines}

Two earlier supervised fine-tuning approaches both plateau at ROUGE-L~$\approx$~0.20 regardless of teacher size or dataset format.
\textit{Basic SFT} trained the student on $\sim$6,000 teacher text responses using cross-entropy loss (teacher: Gemma-2~27B via Ollama).
\textit{Chain-of-thought SFT} added structured \texttt{<thought>} tags using a Gemma-3~27B teacher, which improved response formatting while leaving the performance ceiling unchanged.
The failure mode is the same in both cases: cross-entropy trains the student to predict the single most likely token at each position, discarding the teacher's uncertainty and relative confidence across near-equivalent outputs.
In legal text this loss is costly. Statutory provisions often admit multiple valid framings (for example, a provision may apply under two different acts), and the teacher's soft distribution over output alternatives encodes that ambiguity, information that hard-label training throws away~\cite{hinton2015distillingknowledgeneural,borkar2026memorizationdynamicsknowledgedistillation}.

\subsubsection*{Two-Phase Training}

The student \texttt{google/gemma-2-2b-it} is loaded in 4-bit NF4 quantisation.
A LoRA adapter (rank~32, $\alpha$=64, dropout~0.05) is injected across all seven projection layers ($q$, $k$, $v$, $o$, gate, up, down) in every transformer block, yielding $\sim$8~million trainable parameters ($\approx$0.4\% of total), with an adapter checkpoint of $\sim$112~MB.
The full adapter configuration is given in Table~\ref{tab:lora_config}.

\begin{table}[t]
\centering
\caption{LoRA adapter configuration.}
\label{tab:lora_config}
\begin{tabular}{@{}ll@{}}
\toprule
\textbf{Parameter} & \textbf{Value} \\
\midrule
Base model & \texttt{google/gemma-2-2b-it} \\
PEFT type & LoRA \\
Rank ($r$) & 32 \\
Alpha ($\alpha$) & 64 \\
LoRA scaling ($\alpha / r$) & 2.0 \\
Dropout & 0.05 \\
Bias & none \\
Target modules & $q$, $k$, $v$, $o$, gate, up, down projections \\
Task type & \texttt{CAUSAL\_LM} \\
Trainable parameters & $\sim$8M ($\approx$0.4\% of 2B total) \\
Adapter checkpoint & $\sim$112~MB \\
Training VRAM (peak) & $\approx$8~GB (base 4-bit + adapters BF16) \\
Inference VRAM & $\approx$2~GB \\
\bottomrule
\end{tabular}
\end{table}

Training proceeds in two phases (Table~\ref{tab:training_hparams}).

\textbf{Phase~1} runs for 3~epochs at $lr=10^{-4}$ using cross-entropy loss on assistant response tokens only, via a custom data collator that masks all tokens preceding the Gemma-2 response delimiter.
The purpose is style convergence before distributional supervision begins: starting KL training on a freshly initialised student caused loss oscillation and gradient instability that prevented convergence.
Phase~1 resolves this by first teaching the student the chain-of-thought response format, which gives Phase~2 a stable initialisation point.

\textbf{Phase~2} runs for 3~epochs at $lr=5\times10^{-5}$ with the combined loss:
\begin{equation}
\mathcal{L} = (1 - \lambda)\,\mathcal{L}_{\mathrm{CE}} + \lambda\,\tau^2\,D_{\mathrm{KL}}\!\left(p_T^{(\tau)} \,\|\, p_\theta^{(\tau)}\right)
\label{eq:loss}
\end{equation}
where $\lambda=0.4$ and $\tau=4.0$.
At each token position the student logits are divided by $\tau$ and passed through log-softmax; the result is compared with the stored top-50 teacher distributions using KL divergence with sum reduction.
Both loss terms are averaged across valid sequence positions and combined before backpropagation.
The $\tau^2$ scaling factor in \eqref{eq:loss} compensates for the reduced gradient magnitude of soft targets at high temperature, following Hinton et al.~\cite{hinton2015distillingknowledgeneural}.
A third polishing phase at $lr=10^{-5}$ was attempted and abandoned after it produced a 6.8\% ROUGE-L regression, which confirmed that Phase~2 had already converged.\footnote{Practitioners resuming QLoRA checkpoints: \texttt{PeftModel.from\_pretrained} must be called with \texttt{is\_trainable=True}; the default inference mode silently freezes all LoRA gradients without raising an error.}

\begin{table}[t]
\centering
\caption{Progressive distillation training hyperparameters.}
\label{tab:training_hparams}
\begin{tabular}{@{}lcc@{}}
\toprule
\textbf{Parameter} & \textbf{Phase 1} & \textbf{Phase 2} \\
\midrule
Epochs & 3 & 3 \\
Learning rate & $1\!\times\!10^{-4}$ & $5\!\times\!10^{-5}$ \\
Effective batch size & 16 & 16 \\
Loss & CE only & 0.6\,CE + 0.4\,KL \\
KL temperature $\tau$ & - & 4.0 \\
Top-K teacher logits & - & 50 \\
Base quantisation & 4-bit NF4 & 4-bit NF4 \\
\bottomrule
\end{tabular}
\end{table}

\begin{figure}[t]
    \centering
    \includegraphics[width=\columnwidth]{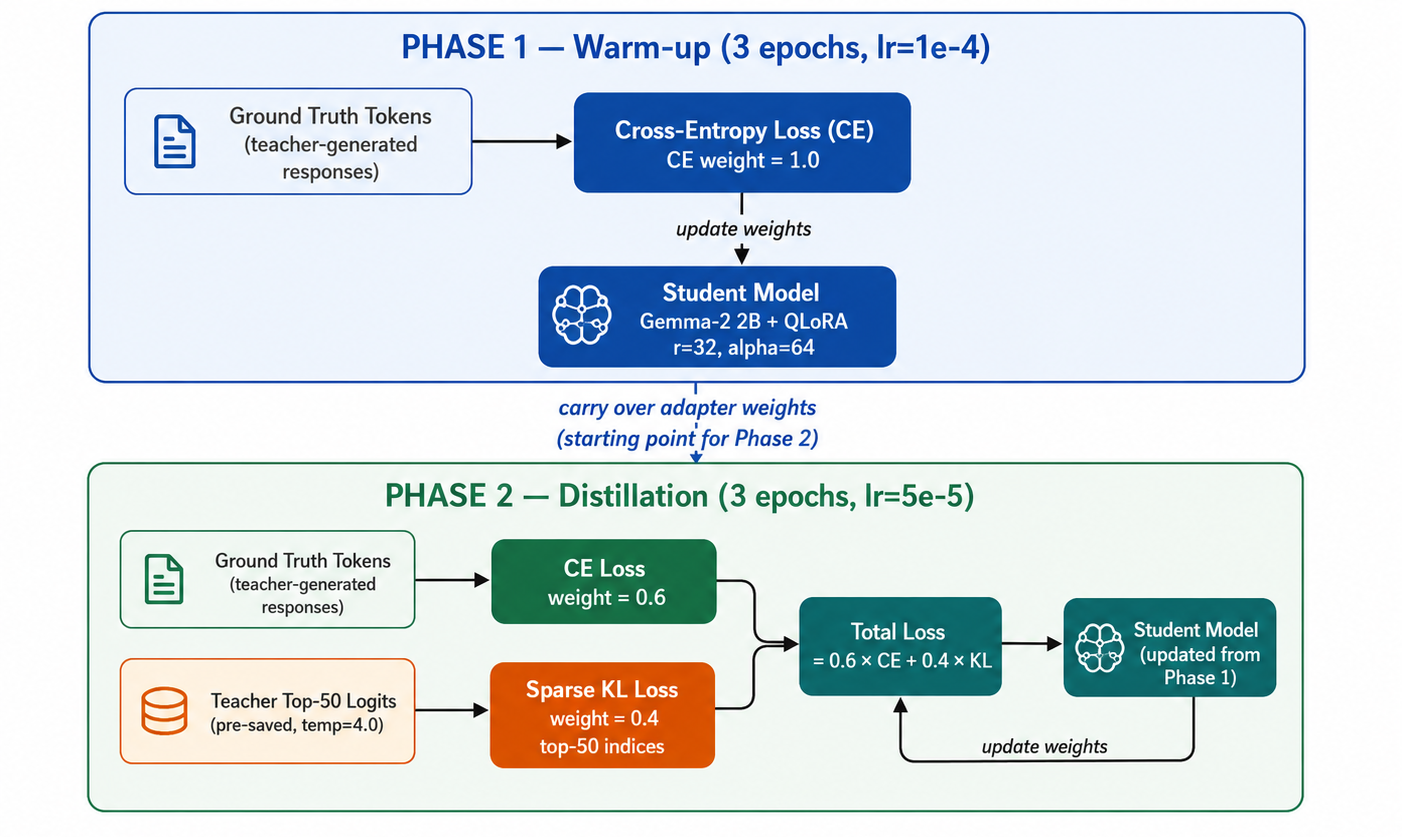}
    \caption{Progressive KD: Phase~1 (CE only) aligns response style; Phase~2 (CE + sparse KL against top-50 teacher logits) transfers the teacher's distributional signal. Phase~3 was disabled after causing regression.}
    \label{fig:progressive_kd}
\end{figure}

\subsection{Hybrid RAG Pipeline}

Each user query is first expanded into three variants, the original plus two paraphrases generated by the 9B teacher via Ollama, to bridge the phrasing gap between everyday legal questions and formal statutory draftspeak.
Each variant is scored against all 36,029~passages as $0.6\times\text{cosine}(q,p) + 0.4\times\text{BM25}(q,p)$.
The 60/40 weighting addresses a structural property of Bangladeshi statutory text: provisions use formal identifiers such as ``section~5 of the Companies Act 1994'' that dense embeddings may not surface from an informal user query, but that BM25 matches precisely through lexical overlap.
Passages with cosine similarity below 0.40 are filtered before re-ranking; this threshold, selected by empirical inspection of retrieval output, removes chunks that match on boilerplate headings (\textit{Definitions}, \textit{Short Title}) without topical relevance to the query.
The top-10 candidates across all query variants are re-ranked by \texttt{ms-marco-MiniLM-L-6-v2}, which scores query--passage pairs jointly rather than independently; the top-3~passages are then formatted with source metadata and injected into the generation prompt.
Table~\ref{tab:rag_comparison} shows the progression from the initial baseline to the final pipeline.

\begin{table}[t]
\centering
\caption{Original vs.\ improved (final) RAG pipeline configuration.}
\label{tab:rag_comparison}
\resizebox{\columnwidth}{!}{%
\begin{tabular}{@{}lll@{}}
\toprule
\textbf{Component} & \textbf{Original} & \textbf{Improved (Final)} \\
\midrule
Query variants & 1 (original only) & 3 (original + 2 paraphrases) \\
Candidates retrieved & 3 & 10 \\
Final chunks returned & 3 & 3 \\
Similarity threshold & None & cosine $\geq$ 0.40 \\
Retrieval method & Semantic only & 60\% semantic + 40\% BM25 \\
Re-ranking model & None & ms-marco-MiniLM-L-6-v2 \\
\bottomrule
\end{tabular}%
}
\end{table}

Each design decision targeted a failure mode observed in the naive baseline.
Semantic-only retrieval missed exact section identifiers, and adding 40\% BM25 recovered them.
Returning only 3 candidates without filtering let in off-topic boilerplate, and the cosine threshold removed it.
Single-query retrieval was brittle to query phrasing, and two paraphrase variants smoothed this sensitivity.
A bi-encoder that selected by score alone sometimes promoted a high-scoring but poorly-matching passage, and cross-encoder re-ranking corrected ordering by evaluating the full query-passage pair jointly.

\begin{figure}[t]
    \centering
    \includegraphics[width=\columnwidth]{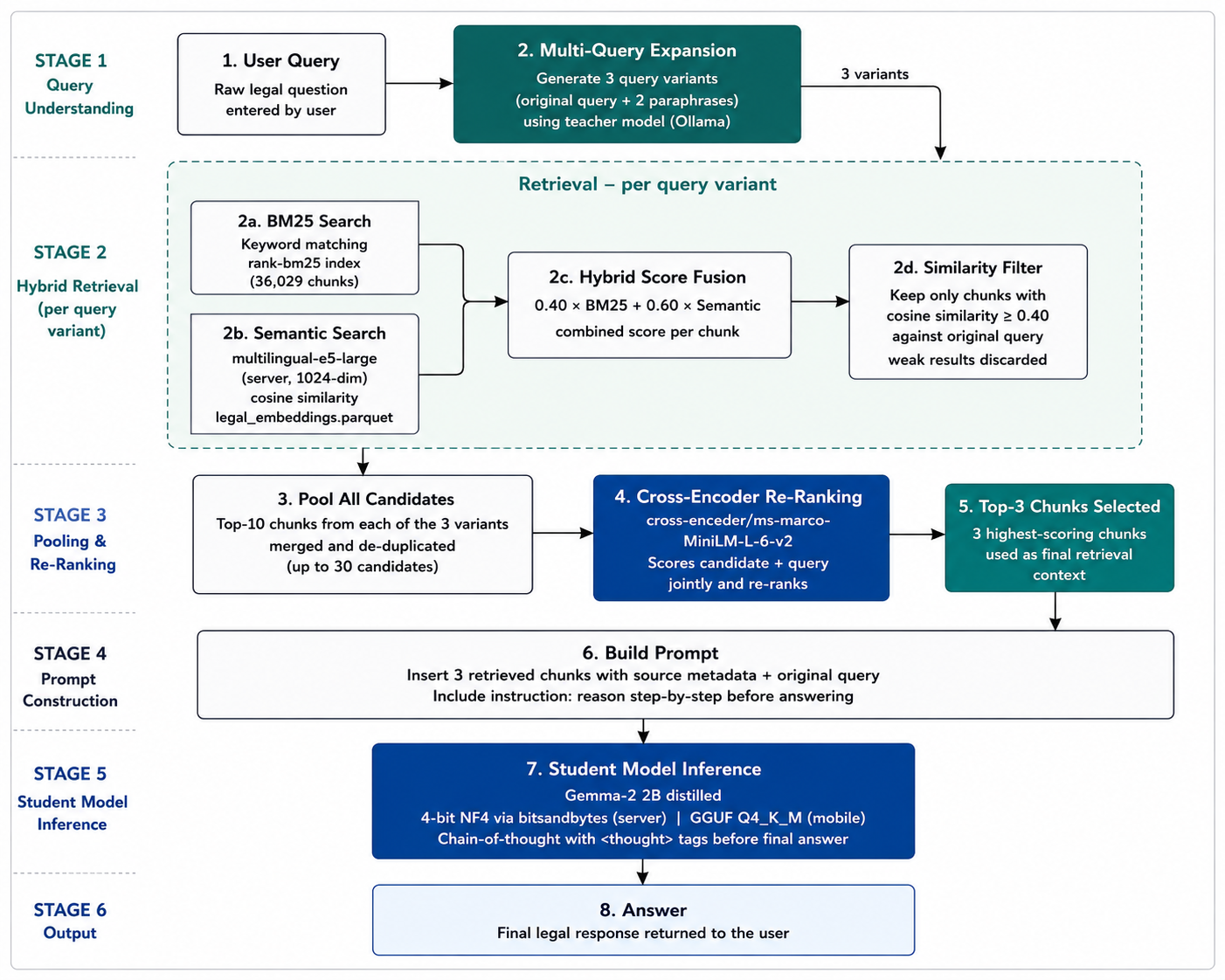}
    \caption{Hybrid RAG pipeline: multi-query expansion generates three query variants; per-variant 60/40 semantic-BM25 scoring, similarity threshold filtering, and cross-encoder re-ranking select the final top-3 passages.}
    \label{fig:rag_pipeline}
\end{figure}

\subsection{Mobile Deployment}

Fully offline operation serves a legal-domain requirement beyond connectivity: citizen queries about rights, labour violations, or criminal procedure should not transit external servers, be logged, or be attributable to a specific user.
The deployment pipeline was designed to keep all computation, all data, and all query text on the user's device.

Conversion proceeds in two steps: the trained LoRA adapter is merged into the base model via \texttt{merge\_and\_unload}, producing a 5.23~GB float16 checkpoint, then quantised to Q4\_K\_M format (1.6~GB) using \texttt{llama-quantize}.
One non-obvious requirement: HuggingFace's Gemma-2 tokenizer save omits the SentencePiece binary \texttt{tokenizer.model} that the llama.cpp converter expects, so this file must be copied manually from the local HuggingFace cache before conversion proceeds.

ThesisApp~v1.0 bundles three artefacts: the 1.6~GB GGUF model, a pre-computed embedding matrix (\texttt{embeddings.bin}, 141~MB), and the full corpus text (\texttt{chunks.json}, 45~MB).
At query time, hybrid retrieval runs entirely on-device without network access.
BM25 scoring uses a Unicode-aware JavaScript inverted index covering both English and Bengali tokens.
Semantic scoring uses \texttt{intfloat/multilingual-e5-large} via ONNX Runtime in a native Kotlin module: the 141~MB embedding matrix is memory-mapped off the application heap, so only a compact score vector crosses the React Native JS bridge per query.
The same 60/40 weighting and cosine threshold ($\geq$0.40) apply; the top-3 retrieved passages are injected into the generation prompt.
Cross-encoder re-ranking and multi-query expansion are omitted, since both require either an additional model loaded in memory or a live Ollama server, neither of which is available in an offline mobile context.
Language model inference runs through \texttt{llama.rn}, a React Native binding for \texttt{llama.cpp}, on the device CPU.

The mobile GGUF benchmark (ROUGE-L~0.4639) was conducted on PC with the lighter \texttt{e5-small} encoder as a conservative proxy for on-device conditions; the deployed application uses e5-large, so 0.4639 is a lower bound on real deployment quality.

\begin{figure}[t]
    \centering
    \includegraphics[width=\columnwidth]{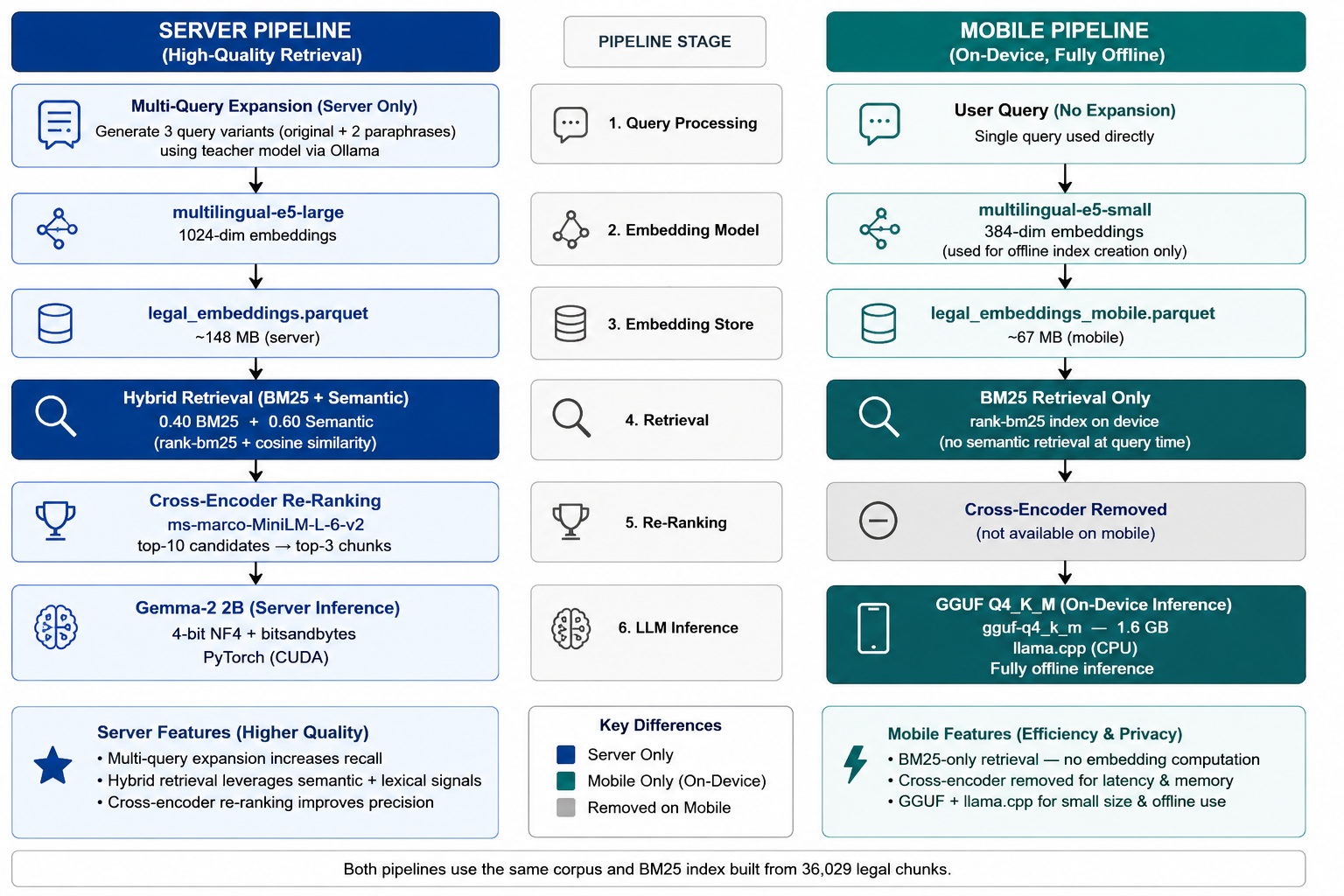}
    \caption{Server vs.\ mobile pipeline comparison: cross-encoder re-ranking and multi-query expansion are removed; on-device hybrid BM25+semantic retrieval is retained via ONNX Runtime (Kotlin native module).}
    \label{fig:mobile_vs_server}
\end{figure}

\section{Experiments and Results}

\subsection{Experimental Setup}

Every stage of the work, including training, dataset generation, evaluation, and GGUF conversion, ran locally on a single NVIDIA RTX 4080 Super (16~GB VRAM) with no cloud compute, which confirms the pipeline runs on consumer hardware.
The evaluation set is a fixed pool of 50~queries drawn from the end of the synthetic query pool to prevent overlap with training data, spanning constitutional law, criminal procedure, property law, and administrative regulations.
Each query was scored under three conditions: the 9B teacher with RAG (the reference), the distilled student without RAG, and the distilled student with RAG.
ROUGE-L, BLEU, and BERTScore~F1 are computed against the teacher-with-RAG reference.

Automated LLM judge scores (faithfulness and relevance, 1--5, from Gemma-3~27B) proved unreliable as primary signals~\cite{li2025llmsreliablyjudgeyet,xu2026positionbiaslljudge}.
For faithfulness, 35 queries scored~1 and 9~scored~5, with only 6~intermediate values (mean 1.94); for relevance, 17~scored~1 and 25~scored~5 (mean 3.46).
The inconsistency is concrete: 15~queries with ROUGE-L above 0.50 received faithfulness scores of~1, a contradiction that exposes the judge's failure to track text-overlap quality~\cite{li2025llmsreliablyjudgeyet}.
This bimodality matches documented failures of small judge models at temperature zero with constrained output length.
Judge scores are reported for completeness but excluded from stage comparisons and ranking decisions.

\subsection{Stage-by-Stage Comparison}

Walking through the configurations (Table~\ref{tab:stage_comparison}) shows where the gains actually come from.
Both SFT approaches plateau at ROUGE-L~$\approx$~0.20 no matter the teacher size or dataset format, which tells us the cross-entropy training signal cannot transfer legal reasoning.
Switching to sparse KL divergence raises ROUGE-L to 0.3935 right away, nearly doubling the SFT ceiling in one change.
After that, dataset expansion (2,806 $\to$ 9,429 samples), hyperparameter tuning ($\lambda$: 0.5 $\to$ 0.4; LoRA rank: 16 $\to$ 32), and corpus re-chunking add up to a final ROUGE-L of 0.4715, a 103\% improvement over the undistilled baseline with retrieval (0.2323).

\textbf{Dataset size.}
Performance went up at every expansion step with no sign of saturation.
From 2,806 to 5,710 samples, ROUGE-L gained +4.3\% and BERTScore +9.3\%.
From 5,710 to 9,429 samples, ROUGE-L gained another +6.6\% and BERTScore +6.3\%.
Returns shrink but stay positive at each step, which suggests the model has not hit a ceiling on this corpus and more SeqKD data would probably keep helping.

\begin{table*}[t]
\centering
\caption{Experimental configurations and results on the 50-query evaluation set with RAG. All KD rows used intfloat/multilingual-e5-large for retrieval; the Mobile GGUF row used e5-small on PC as a conservative proxy.}
\label{tab:stage_comparison}
\resizebox{\textwidth}{!}{%
\begin{tabular}{@{}llcccc@{}}
\toprule
\textbf{Configuration} & \textbf{Training data} & \textbf{LoRA $r$} & \textbf{$\lambda$} & \textbf{ROUGE-L} & \textbf{BERTScore F1} \\
\midrule
No adapter (RAG only)                        & -              & -  & -  & 0.2323          & 0.2340 \\
Text-only SFT                                & $\sim$6k answers & 64   & -  & $\approx$0.20   & -    \\
Chain-of-thought SFT                         & $\sim$6k CoT     & 8    & -  & $\approx$0.20   & -    \\
KD, $\lambda$=0.5                            & 2,806 SeqKD      & 16   & 0.5  & 0.3935          & 0.4720 \\
KD, $\lambda$=0.3                            & 2,806 SeqKD      & 16   & 0.3  & 0.4238          & 0.4890 \\
KD, $r$=32, $\lambda$=0.4                    & 5,710 SeqKD      & 32   & 0.4  & 0.4422          & 0.5343 \\
\textbf{KD final (36,029-chunk corpus)}      & \textbf{9,429 SeqKD} & \textbf{32} & \textbf{0.4} & \textbf{0.4715} & \textbf{0.5679} \\
Mobile GGUF Q4\_K\_M (PC benchmark)         & 9,429 SeqKD      & 32   & 0.4  & 0.4639          & 0.5860 \\
\bottomrule
\end{tabular}%
}
\end{table*}

\begin{figure}[t]
    \centering
    \includegraphics[width=\columnwidth]{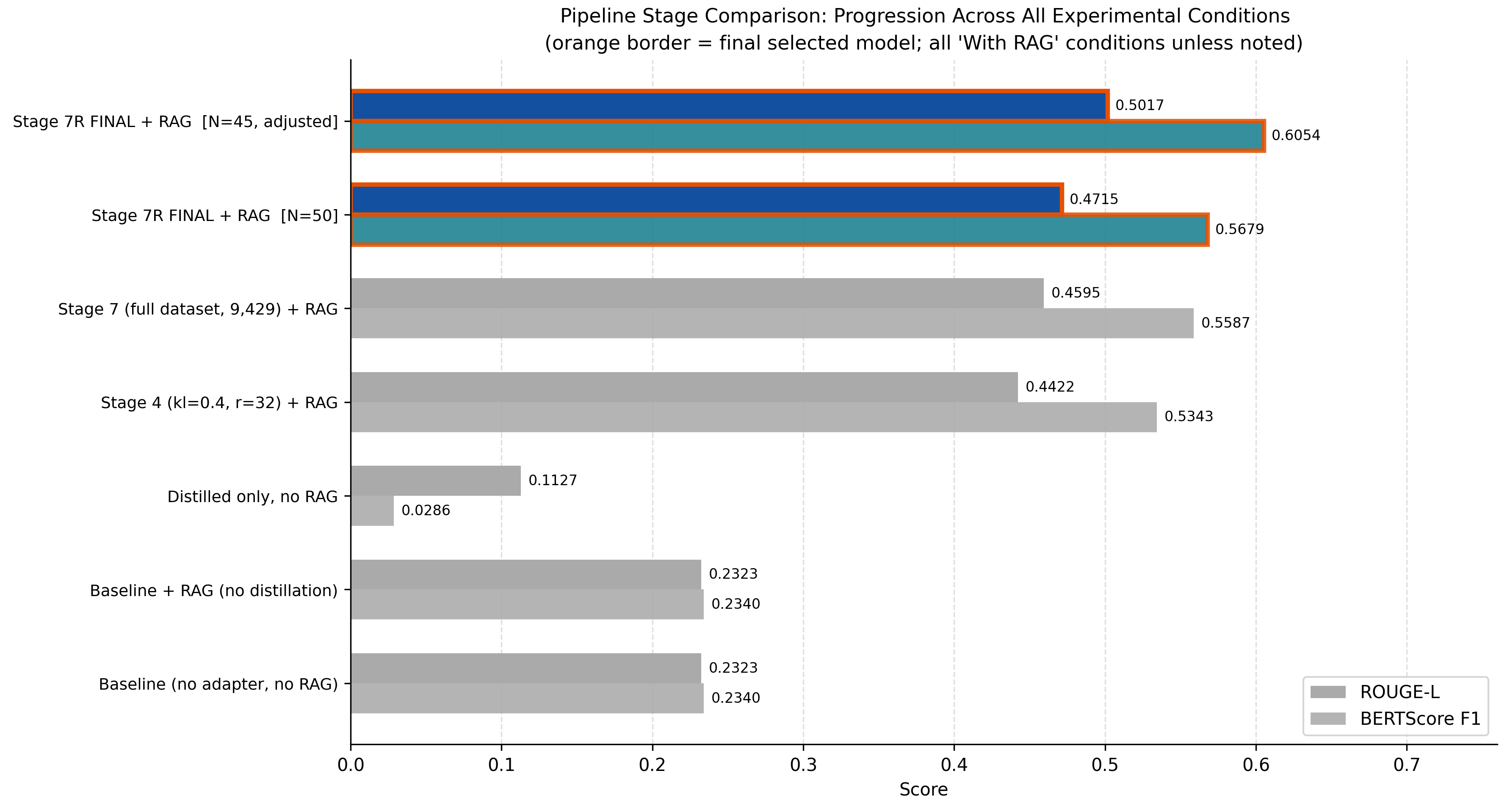}
    \caption{Stage-by-stage ROUGE-L and BERTScore F1 progression. The transition from text-only SFT to sparse KL distillation (Stage~3S1) accounts for the largest single gain.}
    \label{fig:stage_comparison}
\end{figure}

\subsection{Ablation Studies}

\textbf{KL weight.}
$\lambda=0.4$ produced the best combined performance across both primary metrics (Table~\ref{tab:kl_ablation}).
At $\lambda=0.5$ the KL signal dominates and degrades answer coherence: BERTScore falls 8.4\% relative to $\lambda=0.4$, which indicates the student is over-fitted to the teacher's distribution at the cost of fluency.
At $\lambda=0.3$ the distributional signal is too weak to meaningfully exceed cross-entropy alone.

\begin{table}[t]
\centering
\caption{KL weight ablation (5,710-sample dataset, LoRA $r$=32).}
\label{tab:kl_ablation}
\begin{tabular}{@{}cccc@{}}
\toprule
$\lambda$ & \textbf{ROUGE-L} & \textbf{BERTScore} & \textbf{Faithfulness} \\
\midrule
0.3                  & 0.4238 & 0.4890 & 3.46/5 \\
\textbf{0.4 (final)} & \textbf{0.4422} & \textbf{0.5343} & 3.44/5 \\
0.5                  & 0.4324 & 0.4930 & 3.04/5 \\
\bottomrule
\end{tabular}
\end{table}

\textbf{Chunk size and corpus split.}
Chunking strategy had a larger impact than expected (Table~\ref{tab:chunk_ablation}).
A 256-token ceiling bisected Bangladeshi statutory provisions mid-clause, fragmenting the legal reasoning context that the retriever needs to surface a complete provision, a 10.1\% ROUGE-L drop relative to the 450-token improved baseline.
Replacing the original fixed-window splitter with a sentence-boundary-aware splitter, producing 36,029~chunks from the same source documents versus $\sim$8,000, recovered a further 0.012 ROUGE-L by preserving provision boundaries and reducing mid-clause truncation.

\begin{table}[t]
\centering
\caption{Chunk size ablation (9,429-sample dataset, LoRA $r$=32, $\lambda$=0.4).}
\label{tab:chunk_ablation}
\begin{tabular}{@{}lrcc@{}}
\toprule
\textbf{Chunk configuration} & \textbf{Chunks} & \textbf{ROUGE-L} & \textbf{BERTScore} \\
\midrule
450t (original splitter)       & $\sim$8,000 & 0.4595 & 0.5587 \\
\textbf{450t (improved splitter)} & \textbf{36,029} & \textbf{0.4715} & \textbf{0.5679} \\
256t                           & 49,806      & 0.4130 & 0.5271 \\
\bottomrule
\end{tabular}
\end{table}

\subsection{Contribution of RAG vs.\ Distillation}

Ablating each component separately (Table~\ref{tab:rag_contribution}) reveals a striking asymmetry.
The distilled student without RAG (ROUGE-L~0.1120) scores \textit{below} the undistilled baseline with RAG (0.2323): without retrieved context, the student generates fluent but factually ungrounded responses that diverge from the teacher reference more than the raw base model does.
The combined system (0.4715) substantially exceeds both components, which confirms that distillation and RAG are not substitutes; they address different failure modes.
Distillation equips the student with legal reasoning structure: how to analyse a provision, apply it to a query, and state a legal conclusion.
RAG provides factual grounding: which provision applies to this specific query.
Neither alone is sufficient.

\begin{table}[t]
\centering
\caption{Isolation of distillation and RAG contributions (Stage~7R evaluation set).}
\label{tab:rag_contribution}
\begin{tabular}{@{}lccc@{}}
\toprule
\textbf{Configuration} & \textbf{ROUGE-L} & \textbf{BLEU} & \textbf{BERTScore} \\
\midrule
Baseline (no adapter) + RAG   & 0.2323 & 0.0753 & 0.2340 \\
Distilled student, no RAG     & 0.1120 & 0.0208 & 0.0281 \\
\textbf{Distilled student + RAG} & \textbf{0.4715} & \textbf{0.2551} & \textbf{0.5679} \\
\bottomrule
\end{tabular}
\end{table}

\begin{figure}[t]
    \centering
    \includegraphics[width=\columnwidth]{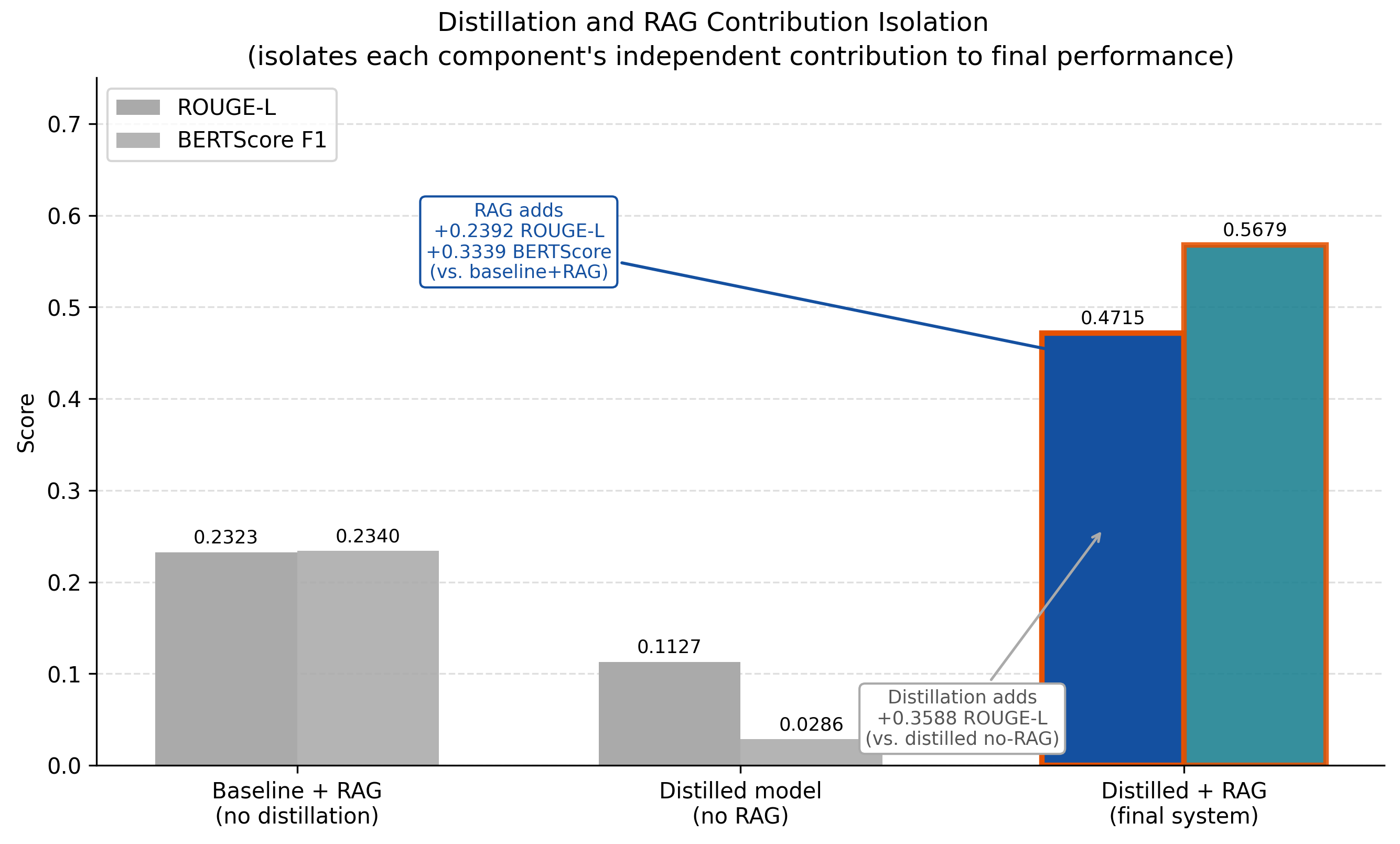}
    \caption{Three-way isolation: undistilled base + RAG (0.2323), distilled student without RAG (0.1120), and distilled student + RAG (0.4715). Neither component alone achieves the combined result.}
    \label{fig:distillation_contribution}
\end{figure}

\subsection{Mobile Deployment Quality}

Q4\_K\_M quantisation keeps primary metric quality within 2\% of the server evaluation (Table~\ref{tab:mobile_quality}).
BERTScore rises 3.2\%, probably because dropping multi-query paraphrase averaging cuts reference-response scoring noise.
Faithfulness improves slightly (+5.2\%, 1.94$\to$2.04/5). This fits with single-query retrieval returning passages matched directly against the user's original phrasing rather than the averaged result of three query variants, which gives the judge simpler, more directly grounded context to verify.
Treat the benchmark figures as a lower bound, since the PC proxy used e5-small whereas the deployed application uses e5-large via ONNX Runtime.

\begin{table}[t]
\centering
\caption{Server (Stage~7R) vs.\ mobile GGUF Q4\_K\_M quality and efficiency.}
\label{tab:mobile_quality}
\begin{tabular}{@{}lccc@{}}
\toprule
\textbf{Metric} & \textbf{Stage 7R} & \textbf{Mobile GGUF} & \textbf{$\Delta$} \\
\midrule
ROUGE-L         & 0.4715 & 0.4639 & $-$1.6\% \\
BLEU            & 0.2551 & 0.2463 & $-$3.4\% \\
BERTScore F1    & 0.5679 & 0.5860 & +3.2\%   \\
Faithfulness    & 1.94/5 & 2.04/5 & +5.2\% \\
Avg.\ latency (s)   & 10.7   & 10.3   & $-$3.7\%  \\
\bottomrule
\end{tabular}
\end{table}

On the Pixel~6 (Android~16, Tensor G1 CPU), the application runs at 4--8~tokens per second, producing 100--200 token responses in 15--40~seconds; first-launch initialisation takes $\approx$30~seconds.
It needs no network access at any stage.

The server inference comparison (Table~\ref{tab:inference_speed}) turned up something we did not expect: RAG \textit{reduces} student response latency instead of increasing it.
Without retrieval context the student generated 411~tokens on average (21.6~s); with RAG, 199~tokens (10.7~s).
Retrieved statutory context gives the student a factual anchor that stops it from generating verbose, speculative completions, and that halves latency as a side effect of improving quality.

\begin{table}[t]
\centering
\caption{Inference speed: teacher vs.\ student on the RTX~4080 Super server.}
\label{tab:inference_speed}
\resizebox{\columnwidth}{!}{%
\begin{tabular}{@{}llll@{}}
\toprule
\textbf{Metric} & \textbf{Teacher (9B)} & \textbf{Student, no RAG} & \textbf{Student + RAG} \\
\midrule
Avg.\ latency (s) & 11.2 & 21.6 & 10.7 \\
Avg.\ tokens generated & $\sim$44 words & 411 tokens & 199 tokens \\
Avg.\ tokens/sec & - & 19.0 & 18.7 \\
VRAM required & $\sim$6~GB & $\sim$2~GB & $\sim$2~GB \\
\bottomrule
\end{tabular}%
}
\end{table}

\subsection{Multilingual (Bangla) Evaluation}

The cross-lingual evaluation handles a practical fact: many Bangladeshi citizens are more comfortable asking legal questions in Bangla than in English, even though the statutory corpus is mostly English.
We evaluated fifty Bangla legal queries in the server configuration. \texttt{intfloat/multilingual-e5-large} maps Bangla and English into a shared embedding space, so retrieval works without language detection or query translation.
Without RAG, ROUGE-L~0.0481 and BERTScore~0.6031 confirm the model cannot generate Bangla legal content from parametric memory alone, since the training data was English-only.
With RAG, ROUGE-L rises to 0.4083 and BERTScore to 0.8133: retrieved English statutory passages give the model the grounding it needs even when the query is in Bangla.
Responses come back in English, which is a real limitation: citizens who cannot read English cannot use the output directly.
Closing that gap would take a dedicated Bangla legal QA training corpus.

\subsection{Human Expert Evaluation}

Automatic metrics measure overlap with a reference text, not legal correctness.
To assess whether responses would be useful to a citizen seeking legal guidance, a practising lawyer rated all 50~Stage~7R responses on a five-point scale covering accuracy, clarity, and practical usefulness (Table~\ref{tab:human_eval}).
90\% of responses received a rating of~4 or~5, with a weighted mean of 4.16/5 (computed as $(5{\times}13 + 4{\times}32 + 3{\times}5)/50$), and no response fell below~3.
The five responses rated~3 correspond exactly to the structurally ambiguous queries identified in Table~\ref{tab:hard_queries}, which contain decontextualised references such as ``this Act'' or multi-domain hypotheticals.
The evaluator independently noted the ambiguous query formulation as the source of the reduced ratings, which confirms these are query-level failures rather than model failures.

\begin{table}[t]
\centering
\caption{Human expert evaluation by a practising lawyer (N=50, Stage~7R + RAG).}
\label{tab:human_eval}
\begin{tabular}{@{}clcc@{}}
\toprule
\textbf{Rating} & \textbf{Description} & \textbf{Count} & \textbf{\%} \\
\midrule
5 & Legally accurate, immediately useful      & 13 & 26 \\
4 & Legally sound, minor omissions            & 32 & 64 \\
3 & Partially correct, lacking detail         &  5 & 10 \\
2 & Significant errors                         &  0 &  0 \\
1 & Factually wrong or harmful                &  0 &  0 \\
\midrule
\multicolumn{2}{l}{\textbf{Weighted mean}} & \multicolumn{2}{l}{\textbf{4.16\,/\,5.00}} \\
\bottomrule
\end{tabular}
\end{table}

\subsection{Hard Query Analysis and Adjusted Scores}

Five queries scored poorly across every training configuration, independent of model quality, which signals that the problem is in the query rather than the model.
All five share the same structural property: they contain underspecified or decontextualised references whose legal intent is ambiguous even to a human reader (Table~\ref{tab:hard_queries}).

\begin{table}[t]
\centering
\caption{Five hard queries: Stage~7R ROUGE-L, BERTScore, and root cause.}
\label{tab:hard_queries}
\resizebox{\columnwidth}{!}{%
\begin{tabular}{@{}lllp{4.5cm}@{}}
\toprule
\textbf{Query} & \textbf{ROUGE-L} & \textbf{BERTScore} & \textbf{Root cause} \\
\midrule
Q38 & 0.152 & 0.155 & Decontextualised: ``this statutory body'' with no named referent \\
Q2  & 0.161 & 0.272 & Decontextualised: ``this Act'' with ambiguous temporal framing \\
Q6  & 0.191 & 0.183 & Speculative: penalty mechanism not stated in corpus \\
Q15 & 0.218 & 0.256 & Multi-domain hypothetical spanning two separate Acts \\
Q4  & 0.280 & 0.289 & Pre-existence hypothetical not covered in corpus \\
\bottomrule
\end{tabular}%
}
\end{table}

The failure mechanism is retrieval ambiguity rather than generation failure.
For an underspecified query, the teacher and student retrieve different statutory passages; both responses may be legally reasonable given their respective retrieved contexts, but they use different statutory text, and ROUGE-L penalises any divergence regardless of individual legal correctness.
So for these five queries the metric measures reference-match rather than legal quality.
The human evaluator reached the same conclusion independently: all five received a rating of~3, and the evaluator explicitly noted ambiguous query formulation as the cause.

Excluding these five queries (N=45) raises all three automatic metrics by a consistent margin:

\begin{table}[t]
\centering
\caption{Full set (N=50) vs.\ adjusted set excluding five hard queries (N=45).}
\label{tab:adjusted_scores}
\begin{tabular}{@{}lccc@{}}
\toprule
\textbf{Metric} & \textbf{N=50} & \textbf{N=45} & \textbf{$\Delta$} \\
\midrule
ROUGE-L      & 0.4715 & 0.5017 & +0.0302 \\
BLEU         & 0.2551 & 0.2783 & +0.0232 \\
BERTScore F1 & 0.5679 & 0.6054 & +0.0375 \\
\bottomrule
\end{tabular}
\end{table}

The $\approx$0.03 ROUGE-L and $\approx$0.04 BERTScore gaps are consistent across all three metrics, which indicates these five queries contribute a fixed, predictable downward bias.
Both the full-set and adjusted scores are reported; the full-set figures are primary to avoid selection bias, and the adjusted figures bound the system's performance on well-formed queries.

\section{Discussion}

\textbf{Why progressive KD outperforms SFT.}
The 0.20 ROUGE-L plateau under SFT is a training signal failure, not a data or capacity failure.
Cross-entropy sees only one correct token per position; the teacher's relative confidence across near-equivalent legal phrasings is discarded.
In a legal domain where multiple statutory provisions may be co-relevant and a good answer often hedges between them, this discarded distributional information is what encodes legal reasoning.
The BERTScore gain from $\lambda=0.3$ to $\lambda=0.4$ (+9.3\%) is substantially larger than the ROUGE-L gain (+0.018), consistent with the KL signal transferring semantic relationships between legal concepts rather than surface text overlap~\cite{borkar2026memorizationdynamicsknowledgedistillation}.

\textbf{Why distillation and RAG are jointly necessary.}
The ablation makes the division of labour explicit: a distilled student without RAG scores \textit{below} the undistilled baseline with RAG (ROUGE-L 0.1120 vs.\ 0.2323).
Distillation teaches the student how to reason over a statutory provision; RAG determines which provision is retrieved.
Neither skill substitutes for the other. A model that can reason well but retrieves the wrong passage will give a legally confident but factually wrong answer, arguably the most dangerous failure mode for a legal AI tool.
Both components must be optimised together.

\textbf{Transferability to other jurisdictions.}
The pipeline ran end-to-end on a single consumer 16~GB GPU.
The inputs it requires, a crawlable statutory repository, a generative teacher model, and an Android build environment, are available for most national legal systems.
The specific techniques (sparse-logit KD, hybrid BM25-semantic retrieval, GGUF mobile deployment) are not Bangladesh-specific, so any jurisdiction with a digital statutory corpus and no existing legal AI could replicate this pipeline with its own legislative texts.

\section{Limitations}

\textbf{Corpus coverage.}
The system only provides for the Constitution and national legislation.
Subordinate legislation or case law are not included and so questions that demand a judicial interpretation cannot be answered with certainty.

\textbf{Bangla response gap.}
While the model can process Bangla queries using cross-lingual retrieval, Bangla legal reasoning is less precise, as the training data was primarily of English.
A corpus of Bangla legal documents would help significantly in improving the quality of the output.

\textbf{LLM judge bimodality.}
The automated judge Gemma-3~27B scored in binary form: 44 of 50 queries were given either~1 or~5 on Faithfulness (35 were given~1 and 9 were given~5), with only~6 scores in between.
ROUGE-L and the ratings of human expert are thus used as primary evaluation signals.

\textbf{Mobile distribution.}
ThesisApp~v1.0 has to be installed through ADB (USB debugging) and isn't available on the Play store at this time.
The production APK signing and store submission are deferred to future tasks.

\textbf{Limited evaluator pool.}
The human expert evaluation was conducted by one practicing lawyer.
Several independent legal validators were recommended to assure inter-rater reliability and minimise evaluator bias, but were not practical in the time and resources of the project.
\section{Conclusion}

We could get responses from a 9B teacher compressed to 2B, running fully offline at 1.6~GB on a Pixel~6 at 4--8~tokens per second, to answer Bangladeshi legal questions with a weghted mean score of 4.16/5.
It took three things to get there: a progressive 2-phase distillation with sparse KL divergence (103\% improvement over the undistilled baseline on ROUGE-L), hybrid BM25-semantic retrieval over 36,029 statutory passages and a full GGUF mobile deployment pipeline in place.
Generalisation across this jurisdiction is enabled by two lessons: the first is that text-only supervised fine-tuning cannot transfer legal reasoning, because it ignores the uncertainty of the teacher's distribution; the second is that retrieval and distillation are two different failure modes, which cannot supercede one another.

There are three directions that are immediately apparent: expansion of the corpus to include case law and subordinate legislation, creating a training set of legal queries in Bangla language so as to allow native-language query responses, and implementation of lightweight query expansion to recover multi-query retrieval without needing a server dependency.

\section*{Acknowledgment}

All praise to the Great Almighty for whom our thesis has been completed without any major interruption. The authors also thank their supervisor Saadat Rafid Ahmed and co-supervisor Dr.\ Farig Yousuf Sadeque for their kind support and advice throughout this work.

\bibliographystyle{IEEEtran}
\bibliography{references}

\end{document}